\documentclass[journal]{IEEEtran}
\usepackage[backend=biber,maxbibnames=1,maxcitenames=1]{biblatex}
\usepackage{hyperref}
\usepackage{svg}
\usepackage{tabularx}
\usepackage{multirow}
\usepackage[flushleft]{threeparttable}
\usepackage{makecell}
\usepackage{url}

\begin{document}
%


\title{
\footnotesize
\framebox[1.01\width]{\parbox{\dimexpr\linewidth-2\fboxsep-2\fboxrule}{If you cite this paper, please use the PEMWN 2023 reference: Z. Huang, K. Zandberg, K. Schleiser, E. Baccelli. U-TOE: Universal TinyML On-board Evaluation Toolkit for Low-Power IoT. In Proc. of 12th IFIP/IEEE PEMWN, Sept. 2023.}}
 \ \\ \ \\ \ \\
\Huge U-TOE: Universal TinyML On-board Evaluation Toolkit for Low-Power IoT}
%
%
%

\author{Zhaolan Huang, Koen Zandberg, Kaspar Schleiser and Emmanuel Baccelli
\thanks{Corresponding Author: Z. Huang. E-Mail: zhaolan.huang@fu-berlin.de}%
\thanks{Z. Huang, K. Zandberg, and K. Schleiser are affiliated with Freie Universität Berlin. 
E. Baccelli is affiliated with Inria, France. 
}
}
%
%

\markboth{.
}%
{Shell \MakeLowercase{\textit{et al.}}: Bare Demo of IEEEtran.cls for IEEE Journals}
%



\maketitle

\begin{abstract}
Results from the TinyML community demonstrate that, it is possible to execute machine learning models directly on the terminals themselves, even if these are small microcontroller-based devices. However, to date, practitioners in the domain lack convenient all-in-one toolkits to help them evaluate the feasibility of executing arbitrary models on arbitrary low-power IoT hardware. To this effect, we present in this paper U-TOE, a universal toolkit we designed to facilitate the task of IoT designers and researchers, by combining functionalities from a low-power embedded OS, a generic model transpiler and compiler, an integrated performance measurement module, and an open-access remote IoT testbed. We provide an open source implementation of U-TOE and we demonstrate its use to experimentally evaluate the performance of various models, on a wide variety of low-power IoT boards, based on popular microcontroller architectures. U-TOE allows easily reproducible and customizable comparative evaluation experiments on a wide variety of IoT hardware all-at-once. The availability of a toolkit such as U-TOE is desirable to accelerate research combining Artificial Intelligence and IoT towards fully exploiting the potential of edge computing.
\end{abstract}


%
\IEEEpeerreviewmaketitle

\section{Introduction}
%
%
%
%
\IEEEPARstart{A}{s} 
Artificial Intelligence (AI) permeates our lives more and more, mechanisms such as deep neural networks~\cite{sze2017dnn} are put to use (or their deployment planned) in more and more places in various distributed systems. In particular, wireless sensor network (WSN) can improve its coverage and connectivity, reduce energy and bandwidth usage by deploying AI onto edge nodes~\cite{electronics10091012}.

The data pipeline with AI typically requires the creation and the use of a \emph{model}, i.e. a layered structure of complex algorithms (also known as \emph{operators}) which interpret data and make decisions based on that data. This model must first be trained (\emph{learning} phase~\cite{sze2017dnn}), before it can be put in production (used for \emph{inference}).

Recent work from the  TinyML~\cite{sanchez2020tinyml,ray2022tinyml-review} community forays into optimizing models to fit tinier resource budgets (and to perform efficiently nevertheless) on low-power microcontrollers in the Internet of Things (IoT). As a consequence, both learning and inference placement possibilities are extended to encompass ultra low-power terminals.

However, generic and convenient open source tools lack for designers tackling a combination of AI and IoT (AIoT), who are required to:
\begin{itemize}
    \item evaluate the performance of their models when placed somewhere along the terminal-edge-cloud continuum, especially when including potential placement on different microcontroller-based devices;
    \item fine-tune their models, and identifying performance bottlenecks at model layer granularity, on different microcontrollers;
    \item select an adequate microcontroller to execute their model, for a targeted task running on a low-power device to-be-designed.
\end{itemize}

This paper thus introduces U-TOE, an open source AIoT toolkit which tightly combines a generic model compiler and a popular low-power IoT operating system to automatically compress, flash and evaluate arbitrary models (output of TensorFlow, PyTorch...) on arbitrary commercial off-the-shelf low-power boards (such as BBC:microbit, nrf52840dk, Arduino Zero, HiFive...) based on the popular microcontroller architectures (ARM Cortex-M, RISC-V, ESP32).

{\bf Paper contributions.} Our contributions are as follows:
\begin{itemize}
    \item We design a universal toolkit for TinyML on-board evaluation (U-TOE). It provides feasibility checks for the deployment of arbitrary model on a given IoT hardware platform. It allows researchers and developers to locate the performance bottleneck of a given model on a target device. The evaluation results enable co-design with other components at system level, help optimizing ML models and configurations for specific use cases, allowing to achieve the best possible performance on target devices.
    
    \item We released the code\footnote{see \url{https://github.com/zhaolanhuang/U-TOE/}} of U-TOE under an open source licence. This implementation enables compilation, flashing, and evaluation of neural network (computational graph based) models from mainstream ML framework onto various low-power boards based on popular ISA.

    \item We provide benchmarks and a comparative experimental evaluation using U-TOE, reproducible both on an open-access IoT testbed and on personal workstations, which provide insights on inference performance with different models on different low-power hardware and
    demonstrate how U-TOE can be re-used by TinyML experimental researchers and developers to fine-tune IoT configurations. 
\end{itemize}
\section{Related work}
Recent work has surveyed \cite{saha2022tinyml-review} the scope of ML frameworks, tasks, metrics, including a comprehensive review on TinyML stack and deployment pipeline. A number of challenges need to be met in order to fit the tiny resource budgets typical of microcontrollers (kiloBytes of memory, power consumption in mWatt, CPU frequency in MHz) while maintaining performance at an acceptable level and retaining portability to extremely polymorphic hardware in this category.

{\bf Embedded IoT Software Platforms --}
Various open source IoT operating systems are used to provide hardware abstraction, and resource sharing primitives as well as convenient peripheral access (e.g. sensor/actuator, network subsystem) on heterogeneous low-power microcontrollers. Prior work such as \cite{hahm2015operating} surveys such operating systems, among which prominent examples include RIOT\cite{baccelli2018riot} and Zephyr\cite{zephyr}. However, such software platforms so far offer very limited support for ML frameworks -- if at all. Moreover, the most advanced support so far are typically hardware- or vendor-specific e.g. with libraries provided by STM32CubeMx or ARM CMSIS-NN.

{\bf Low-power IoT Testbeds --}
Various testbeds offer remote access to fleets of reprogrammable microcontroller-based devices. Prior work such as \cite{lima2019testbeds,mitton2011testbeds} survey such testbeds, among which prominent examples include the open access facility IoT-lab \cite{adjih2015iot-lab}, which offers remote bare-metal access (serial over TCP) to a fleet composed of hundreds of popular low-power boards of various kind.
 
{\bf Benchmarking Suites for TinyML --}
Benchmarking ML on low-power hardware entails a number of challenges \cite{banbury2020benchmarking}. Prior work such as MLPerf Tiny \cite{banbury2021mlperf} provides a standard benchmark suite (a fixed set of representative ML tasks) for evaluating the performance of given hardware, and an online platform for manufacturers to publish their comparative benchmark results. 
In contrast, U-TOE offers a more powerful and more customizable toolkit for performing feasibility's check of user-defined machines learning models on low-power devices, with a greater degree of flexibility and customization.
 
{\bf TinyML Benchmarks --} 
Prior work such as \cite{osman2022tinyml-bench} focuses on performance comparison of different machine learning frameworks on two COTS low-power boards (Arduino Nano BLE 33 and STM32 NUCLEO-F401RE). In particular, it benchmarked two TinyML frameworks, TFLM and X-CUBE-AI over gesture recognition and wake word spotting. Other work such as \cite{sudharsan2021tinyml-bench} tested TFLM models on several microcontroller-based boards. While such papers provide a performance comparison of specific frameworks on specific boards for specific tasks, U-TOE offers greater flexibility and generality, allowing developers to evaluate a wider range of (user-specified) models on a larger variety of low-power devices, and to dive into the execution details of ML models.
 
{\bf TinyML Model Transpiler \& Compilers --}
Compilers such as TVM \cite{chen2018tvm} can be used to automate the transpilation and compilation of models provided by major ML frameworks (TFML, Pytorch etc.) so as to expose low-level routines and optimize them for execution on specific processing unit characteristics (CPU, GPU etc.). An extension of TVM called uTVM was recently introduced, adding smaller hardware targets including a variety of MCUs (microcontroller units).

{\bf TinyML Model Profilers --}
ML-EXray\cite{qiu2022ml-xray} enables TinyML developers to gain visibility into the layer-level details of ML execution and diagnose cloud-to-edge deployment issues. Developers can analyze and debug edge deployment pipelines with high usability, using less than 15 lines of code for fully examination. However, the reliance on Tensorflow Lite restricts the capability to accommodate models from further ML frameworks and deploy on low-power devices.
Major ML frameworks (TFML, Pytorch and MXNet etc.) provide internal profiler \cite{2023_yousefzadeh-asl-miandoab_ProfilingMonitoringDeep}. Such tools allows developers to measure the performance of their models. They can be used to collect metrics such as inference time and memory usage, which can then be analyzed to optimize the model's performance. Though it can provide us execution details in layer level, it still lacks the supports for on-device deployment and evaluation on various IoT devices, while U-TOE is a more general-purpose toolkit that provides a comprehensive solution on a wide range of low-power devices. The above is summarized in Table \ref{tab:related-work-summary}.

\begin{table}[!ht]
\begin{threeparttable}

    \centering
    \caption{Comparison of AIoT frameworks and toolkits.}
    \label{tab:related-work-summary}
    \begin{tabular}{llllll}
    \hline
        Framework & MCU & \makecell[l]{Model\\Type} & \makecell[l]{Remote\\Eval.} & Granularity & \makecell{Model to\\Board Sol.} \\ \hline
        MLPerf & Yes & \makecell[l]{Specified} & No & Model & No \\ 
        ML-EXray & No & TFLite & No & Layer & No \\ 
        TFLite & Yes & TFLite & No & Layer & No \\ 
        Pytorch & No & Torch & No & Layer & No \\ 
        uTVM  & Yes & Universal & No & Operator & No \\
        \bf U-TOE & Yes & Universal & Yes & Operator & Yes\\ \hline
    \end{tabular}
\begin{tablenotes}
    \item Eval. -- Evaluation, Sol. -- Solution
\end{tablenotes}
\end{threeparttable}
\end{table}

\section{Background on TinyML Performance Analysis}

One the one hand, as the most immediate limiting resource budget on microcontrollers concern memory limitations, typically in the order of kiloBytes, TinyML performance evaluation typically focuses primarily on metrics measuring memory consumption -- while keeping an eye on execution speed -- as described below.
On the other hand, TinyML performance analysis can be tackled at different granularity levels: at the global model level, or at the operator level, for finer granularity, as described in the following.

\subsection{Performance Metrics}
The considered metrics offer insights into the feasibility, efficiency and resource utilization of offloading model inference burden to low-power devices. By analyzing these metrics, users can make initial decisions regarding model  selection, optimization techniques, and hardware configurations to maximize performance and minimize the resource footprint on low-power devices.

{\bf Memory (RAM) Consumption --}
This metric measures the amount of dynamic memory space (primary RAM) consumed by the model during inference. It reflects the memory footprint of the model activation and is important for low-power devices that have limited memory resources. Efficient memory utilization allows for the deployment of larger and more complex models on such devices.

{\bf Storage (Flash memory) Consumption --}
This metric quantifies the amount of storage space, typically in terms of Flash memory region, required to store the compute instruction and associated parameters. It reflects the model's storage footprint on the low-power device. Minimizing storage consumption allows for accommodating multiple models on the device or orchestrating with other essential applications.

{\bf Computational Latency --} This metric measures the time consumption  of performing inference for each input sample, either at the model level or at the level of individual operators within the model. It reflects the inference speed of the model on the low-power device and plays crucial role in real-time or latency-sensitive applications. Core clock frequency, cache strategies and communication latency between memory and working core have great impact on this indicator.

{\bf SoC Price --} This metric considers the cost of the System-on-a-Chip (SoC) used in the low-power device. The price of the SoC affects the overall affordability and feasibility of deploying model in large-scale distributed system. Lower-cost SoCs can make the deployment more accessible and cost-effective.

\subsection{Measurement Granularity}
As for performance analysis of machine learning in other domains, TinyML performance can be measured at different granularity levels:

{\bf Per-Model Evaluation --}
At this coarse level, one measures performance of the model as a whole, i.e. the resource footprint incurred by the execution of the model including all its layers and operators. For example, this allows for evaluating the resource consumption for inference with the production-ready code, on a particular industrial hardware setup.

{\bf Per-Operator Evaluation --}
At this level, one measures separately the performance of one or more operators (i.e. one or more components of the model). This per-operator measurement can help identifying specific operators that contribute to performance or inefficiencies, in optimizing the model's efficiency and spotting potential bottlenecks.

\section{U-TOE Toolkit Design \& Implementation}

\subsection{Toolkit Architectural Design}
\begin{figure}[ht]
    \centering
    \includegraphics[width=0.8\columnwidth]{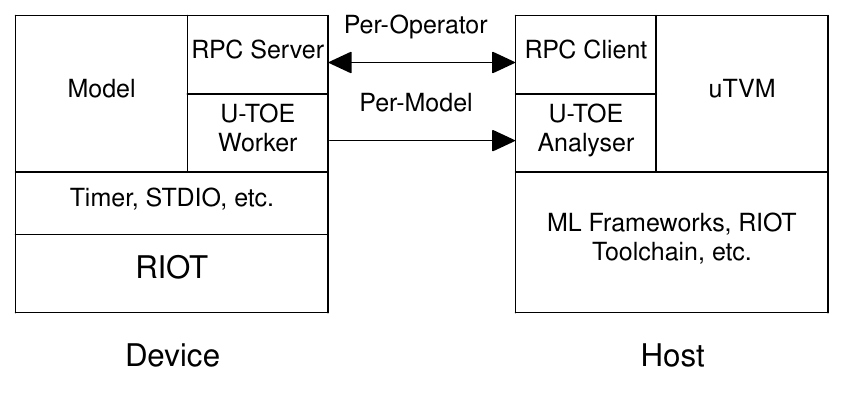}
    \caption{Software architecture and components of U-TOE.}
    \label{fig:utoe-sw-setup}
\end{figure}

U-TOE integrates uTVM and RIOT to perform model compilation, flashing, and evaluation of arbitrary models from mainstream ML frameworks onto various low-power boards. The toolkit is composed of the following key components:
\begin{itemize}
    \item Model Compiler. U-TOE leverages the uTVM compiler to convert arbitrary neural network models into efficient C code. This compiler enhances the efficiency of the models and enables them to be run on low-power devices.

    \item RPC Mechanism. To evaluate the resource consumption at operator level, U-TOE utilizes the Remote Procedure Call (RPC) mechanism of uTVM. The RPC mechanism enables to upload and launch functions on to IoT boards over serial. This is useful for remote testing and profiling, enabling U-TOE to wrap the model operators for measuring the computational latency and memory usage. It is composed of a client on the host and a server on the target device, receiving commands and executable instructions from the host.

    \item OS Environment. RIOT was chosen to provide a lightweight runtime environment for model execution and evaluation on microcontrollers, with its advantages in extensibility and wide-spectrum support for low-power boards.

    \item Evaluation Module. It contains two units: measurement worker and analyser. As shown in Fig.~\ref{fig:utoe-sw-setup}, the measurement worker is deployed on the MCU for acquiring performance metrics at model or operator level. Besides carrying out the measurement of resource footprint, It is in charge of the randomization of model input, and responsible to report metrics data to host device. The analyser runs on the host, statisticizes the uploaded metrics from device and provides human-readable frontend for users. 
\end{itemize}

\begin{figure}[ht]
    \centering
    \includegraphics[width=0.8\columnwidth]{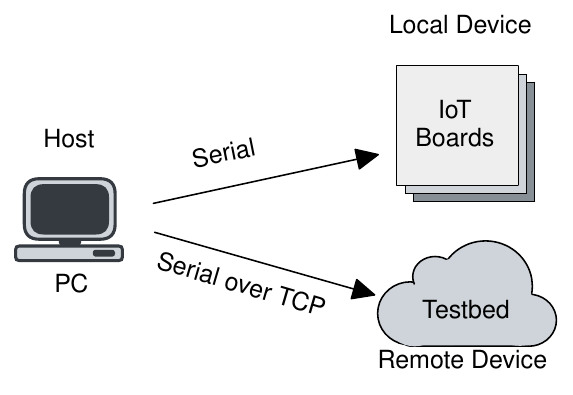}
    \caption{Hardware setup of U-TOE. Users can connect local boards with host PC via serial, or use remote board service on IoT-Testbed.}
    \label{fig:utoe-hw-setup}
\end{figure}

Additionally, U-TOE provides a \emph{connector} for cloud-based IoT testbed which enables seamless interaction with remote boards using serial over TCP. As depicted in Fig.~\ref{fig:utoe-hw-setup}, users can evaluate and benchmark their models on local device, or on a remote, scalable testbed with wide-range of MCUs via U-TOE connector.

\begin{figure*}[ht]
    \centering
    \includegraphics[width=0.65\textwidth]{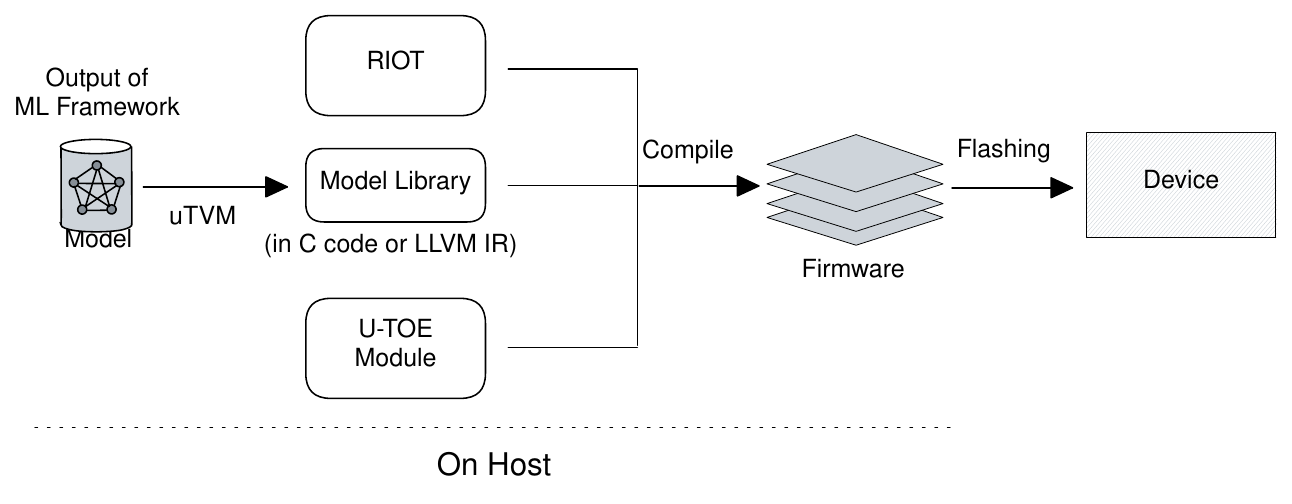}
    \caption{Compilation and deployment workflow of U-TOE. uTVM optimizes and translates model from mainstream ML framework into model library, which is co-compiled and flashed with RIOT and U-TOE components onto target boards.}
    \label{fig:utoe-compile-and-deploy}
\end{figure*}

{\bf Compilation and Deployment --} U-TOE first gathers the specification of target device to decide the compilation options for uTVM and RIOT. Then, as presented in Fig. \ref{fig:utoe-compile-and-deploy}, uTVM generates sub-optimal model library of the input model without scheduling\footnote{A schedule specifies low-level optimization for loop execution, enhancing cache hit and memory access. The optimal one is co-determined by model and device specification, and identified by heuristic search algorithms based on measurements on device. \cite{chen2018tvm} }. Some static optimization strategies are applied in this stage according to target device type. This uTVM-generated library is jointly compiled with RIOT, RPC server and measurement worker into executable firmware, which is then automatically flashed on the device.

{\bf Evaluation --} After deploying the program on the target device, a bidirectional channel -- in Per-Operator evaluation --, or a unidirectional channel -- in Per-Model evaluation -- is set up between host and device, as depicted in Fig.~\ref{fig:utoe-sw-setup}. The measurement worker starts collecting performance metrics at user-specified level and uploads metrics data to analyser. Eventually, users can obtain the overall statistics of model metrics, or catch the performance bottleneck with execution details of each operator. All the raw metrics data are saved in log file for further, user-customized analysis.

\subsection{Measurement Procedure}
We designed two measurement procedures to support evaluation at different granularity. The procedures run across multiple components of the toolkit and the most workloads are primary on the target board. Following steps describes the measurement routine after compilation of executable program. The steps marked with \textbf{bold number} are executed on the target board.

\newcommand{\itembf}{\item[\stepcounter{enumi}\textbf{\arabic{enumi})}]}
{\bf Per-Model Evaluation --}
This mode focuses on the model performance in actual production environment. Here is the corresponding measurement routine:
\begin{enumerate}[]
    \item Calculate model memory and storage consumption based on ELF file. We disable dynamic memory allocation to enable static analysis of memory footprint.
    \item Deploy executable program to local or remote IoT board.
    \itembf Repeat model inference based on the user-specific number of trials with randomized input on uniform distribution.
    \itembf Record computational latency of each trial.
    \itembf Upload records to host device for further analysis and archive.
\end{enumerate}

At the end of evaluation, results with statistics (e.g. 95\% confidence interval, median, maximum and minimum) are presented on the host device, including computational latency and consumption of memory and storage.

{\bf Per-Operator Evaluation --}
In contrast to per-model evaluation, this mode focuses on the efficiency and resource footprint of each operator, enabling to discover the performance bottleneck inside models. Thanks to the time evaluator inside the uTVM's RPC mechanism, we can measure computational latency at operator level. The high abstraction of timer and serial in RIOT allows us to unified the implementation of time measurement and RPC communication on arbitrary IoT boards. Here is the corresponding measurement routine:

\begin{enumerate}
    \item Analysis memory footprint at operator level utilizing internal API of TVM.
    \item Deploy executable program to local or remote IoT board.
    \itembf Start RPC Server on IoT board.
    \itembf Launch RPC client to benchmark and record execution performance of each operator.
\end{enumerate}

It is noted that the operator structure constructed by uTVM are usually inconsistent with the hand-crafted version in ML framework. That's because uTVM as model compiler applies model optimization (i.e., operator fusion) and inserts execution details (i.e., quantization arithmetic) during convert and compilation, which potentially merge multiple operator into single one or insert additional operators. Nevertheless, 
we annotate the operators from uTVM with model parameters (weights, biases etc.), so that users can associate a specific operator to the corresponding layer.

\section{Experiments Using U-TOE}
We conducted experiments to validate the functionality and compatibility of U-TOE on model side (universal support for model structure and ML frameworks) and on device side (wide-spectrum support for IoT devices). Hence, we dived into two orthogonal directions: For device support, we evaluated a quantized LetNet-5 on various IoT boards; For model compatibility, we evaluated multiple models on a local STM32F746G discovery board. 
\begin{itemize}
    \item \textbf{Model Selection} We selected pre-trained, quantized models from open source repositories\footnote{see \url{https://github.com/ARM-software/ML-zoo} and \url{https://github.com/mlcommons/tiny}}, which target on typical TinyML tasks (Image Classification, Key Word Spotting, Visual Word Wake, Noise Suppression and Abnormal Detection). The weights and activations of the model were quantized to 8-bit integer, yet the inputs and outputs remain in IEEE FP32 format.
    \item \textbf{Model Optimization} We only used built-in, rule-based optimization in uTVM. Thus, all heuristic optimization strategies like model scheduling were disabled.
    \item \textbf{MCU Configuration} We disabled data and instruction cache to observe the "memory wall" effect in ML model. The core clock frequency was pre-set by CPU initialization code in RIOT and is presented with experiment results in Section ~\ref{sec:results}. 
    \item \textbf{Hybrid Deployment} The experiments were conducted both on local and remote IoT boards provided by FIT-IoT Lab. 
\end{itemize}
It is noted that for each evaluation we preset the number of trials to ten in order to address random error.

\section{Analysing U-TOE Measurements}
\label{sec:results}

{\bf Per-Model Evaluation --}
Table \ref{tab:per-model-eval-result-multiboard} presents the resource consumption of LeNet-5 model on various IoT boards, generated by Per-Model evaluation. ARM Cortex-M series MCUs show no significant difference on memory and storage usage, and the computational latency declines as the core frequency increases. Benefits from fully support of DSP and Thumb-2 instruction set, Cortex-M3, -M4 MCUs perform better than Cortex-M0+ with the same core clock frequency. An \textbf{outlier} was discovered on SiFive RISC-V MCU. With the highest core clock frequency this won the least favorable ranking on computational latency and the memory usage. The SiFive RISC-V MCU uses an external, SPI NOR flash for data and program storage, causing a huge performance regression while we disabled the cache.

Table \ref{tab:per-model-eval-result-multimodel} presents the results of various ML models on representative TinyML tasks on single IoT board, proving the universal support of ML framework and model structure. Except for LeNet-5 trained on local host device with Pytorch, all the others came from open source model zoo. Memory and storage columns refer to the resource consumption.

\begin{table*}[ht]
\centering
\begin{threeparttable}
\caption{Evaluation of LeNet-5 model on various IoT boards.}
\label{tab:per-model-eval-result-multiboard}
\centering
\begin{tabular}{llrrlrrr}
\hline
\multirow{2}{*}{Board   / MCU} & \multirow{2}{*}{Core} & \multirow{2}{*}{Memory (KB)} & \multirow{2}{*}{Storage (KB)} & \multicolumn{4}{l}{Computational Latency   (ms)}    \\ \cline{5-8} 
                               &                       &                              &                               & 95\%-CI               & Median  & Min.    & Max. \\\hline
b-l072z-lrwan1 / STM32L072CZ & M0+ @ 32 MHz          & 11.288                       & 64.340                         & {[}261.829, 262.249{]} & 262.187 & 261.350  & 262.216 \\
samr30-xpro / ATSAMR30G18A & M0+ @ 48 MHz          & 11.208                       & 65.168                        & {[}176.936, 176.965{]} & 176.958 & 176.924 & 176.975 \\
arduino-zero   / ATSAMD21G18        & M0+ @ 48 MHz          & 11.292                       & 64.940                         & {[}182.061, 182.082{]} & 182.068 & 182.051 & 182.098 \\
rpi-pico   / RP2040        & M0+ @ 125 MHz   &   28.704     &    109.504  &  [70.108, 70.130]  &   70.117  &   70.091 & 70.151 \\

openmote-b   / CC2538SF53          & M3 @ 32 MHz           & 11.100                         & 66.080                         & {[}200.337, 200.384{]} & 200.367 & 200.323 & 200.404 \\
IoT-LAB M3 / STM32F103REY    & M3 @ 72 MHz           & 11.296                       & 62.260                         & {[}97.740, 97.757{]}    & 97.751  & 97.733  & 97.764  \\

nucleo-wl55jc   / STM32WL55JC  & M4 @ 48 MHz           & 11.288                       & 63.180                         & {[}98.649, 98.668{]}   & 98.661  & 98.637  & 98.679  \\
nrf52840dk   / nRF52840        & M4 @ 64 MHz           & 11.348                       & 61.332                        & {[}66.078, 66.112{]}   & 66.088  & 66.087  & 66.163  \\
b-l475e-iot01a   / STM32L475VG & M4 @ 80 MHz           & 11.288                       & 61.604                        & {[}52.900, 52.901{]}     & 52.901  & 52.900    & 52.902  \\
stm32f746g-disco / STM32F746NG       & M7 @ 216 MHz          & 11.076                       & 64.712                        & {[}39.600, 39.602{]}     & 39.601  & 39.599  & 39.604 \\\hline

esp32-wroom-32 / ESP32-D0WDQ6  & ESP32 @ 80 MHz       & 115.958                       & 157.719                        & {[}85.580, 85.583{]} & 85.582  & 85.576 & 85.584
\\
esp32c3-devkit / ESP32-C3FN4 &  RISC-V @ 80 MHz      &    258.874       &    222.272   & [54.947, 54.957]   &      54.953    &   54.938   &    54.961 \\
sipeed-longan-nano / GD32VF103CBT6  & RISC-V @ 108 MHz &     103.108     &    106.422 & [37.783, 37.789]   &   37.789   &    37.779    &   37.791 \\
hifive1b / SiFive FE310-G002  & RISC-V @ 320 MHz       & 60.884                       & 66.492                        & {[}153.621, 154.166{]} & 153.747 & 153.717 & 154.938 \\
\hline
\end{tabular}
\end{threeparttable}
\end{table*}

\begin{table*}[ht]
\centering
\begin{threeparttable}
\caption{Evaluation of various quantized models on stm32f746-disco board.}
\label{tab:per-model-eval-result-multimodel}
\begin{tabular}{llrrlrrr}
\hline
\multirow{2}{*}{Model} & \multirow{2}{*}{Task} & \multirow{2}{*}{Memory (KB)} & \multirow{2}{*}{Storage (KB)} & \multicolumn{4}{l}{Computational Latency   (ms)} \\ \cline{5-8}
                       &                       &                              &                               & 95\%-CI      & Median      & Min.     & Max.     \\\hline
DS-CNN   Small INT8\tnote{**}    & Keyword Spotting      &    68.992       &    71.796         &   [461.395, 461.396]  &    461.396         &   461.396       &   461.397       \\
MobileNetV1-0.25x INT8\tnote{*}         & Visual Wake Words     &  185.352      &    491.668         &  [1435.937, 1435.938]  &   1435.938           &  1435.938         &    1435.939      \\
LeNet-5 INT8          & Image Classification  &     12.068      &    65.851             &    [39.599, 39.603]          &   39.601   &   39.598       &   39.605     \\
Deep AutoEncoder INT8\tnote{*}         & Anomaly Detection     &    6.532     &   292.696 & [35.637, 35.638]    &     35.638   &    35.638   &    35.639   \\
RNNoise INT8\tnote{**}           & Noise Suppression     &   4.688       &    119.652        &   [12.151, 12.157]    &   12.154   &   12.148       &   12.160      \\\hline
\end{tabular}
\begin{tablenotes}
\item [*] These models originate from MLPerf Tiny Benchmarks repository on \url{https://github.com/mlcommons/tiny}.
\item [**] These models originate from ARM Model Zoo on \url{https://github.com/ARM-software/ML-zoo}.
\item All models were pre-trained and quantized by TFLite, except LeNet-5 INT8 was by Pytorch.
\end{tablenotes}
\end{threeparttable}
\end{table*}

{\bf Per-Operator Evaluation --}
We here used a tiny model with only three layers from TFlite as an example to avoid unnecessary complexity in demonstration, with output results presented in Table \ref{tab:per-ops-eval-result}. The computational bottlenecks are located in operator \emph{add\_nn\_relu} and \emph{add\_nn\_relu\_1}, and with the highest memory and storage consumption as well. We can trace down the corresponding layers of original model with the hints of associated parameters, which are the weights, bias or other trainable parameters of the model, making it possible to apply optimization strategies on well-targeted layers. 

\begin{table}[h]
\centering
\begin{threeparttable}
\caption{Per-Operator Evaluation Output of TFlite sinus model on stm32f746-disco board.}
\label{tab:per-ops-eval-result}
\begin{tabular}{lrrlll}
\hline
Operators & Time (us) & Time (\%) & Params  & Memory & Storage \\\hline
add\_nn\_relu    & 8.856        & 15.22\%   & p0, p1                 & 0.128       & 0.128        \\
add\_nn\_relu\_1 & 46.682        & 80.23\%   & p2, p3                             & 0.128       & 1.088        \\
add              & 2.646         & 4.54\%   & p4, p5 & 0.068       & 0.068       \\\hline
\end{tabular}
\begin{tablenotes}
    \item The uTVM auto-generated prefix \textit{tvmgen\_default\_fused\_nn\_dense\_} of operator name is not presented for the purpose of clarity. Memory and storage consumption are presented in KB.
\end{tablenotes}
\end{threeparttable}
\end{table}

\section{Reproducible \& Custom U-TOE Experiments}
We released the full source code of the U-TOE toolkit on Github at \url{https://github.com/zhaolanhuang/U-TOE/} under an open source LGPL v3 license. For further details on how to start with U-TOE hands-on, the reader is referred to the comprehensive \textit{Readme.md} in the repository. 


On the one hand, researchers and practitioners who possess IoT hardware that is supported by the open source operating system RIOT (currently 250+ types of boards, using 60+ types of CPUs~\footnote{see \url{https://github.com/RIOT-OS/RIOT/tree/master/boards}} can use U-TOE out-of-the-box, directly on their boards. 

On the other hand, combined with the use of the free open-access testbed IoT-Lab~\footnote{see \url{https://www.iot-lab.info/}}, even researchers and practitioners who do not have such hardware on premises can conduct large scale experimental evaluation campaigns using U-TOE.

{\bf Perspectives --} As RIOT board and CPU support expands and improves over time, and as uTVM also expands support to other architectures in parallel (both open source communities are very active) U-TOE can in a very short time expand its support for new use cases, automatically adding the support of uTVM for new boards, and the support of RIOT for new models. As such U-TOE may oragnically grow and become a useful link between the two communities.

Moreover, while the work on U-TOE in this paper has been focused on inference only on single-core microcontrollers, there is strong potential to extend the toolkit provided by U-TOE to support on-device learning scenarios, and for optimizing exploitation of multi-core microcontrollers.

\section{Conclusion}
In this paper we presented U-TOE, a novel toolkit we designed to substantially facilitate the task of AIoT practitioners, by enabling universal TinML model evaluation. Using U-TOE, the output of arbitrary machine learning frameworks can be evaluated on arbitrary low-power hardware based on different microcontroller architectures, all at once. The wide availability of such a toolkit is indeed desirable to accelerate the field of AIoT. We thus provided a highly re-usable, documented and customisable open source implementation of U-TOE, jointly harnessing the active open source communities around RIOT and uTVM. Finally, we demonstrate the use of U-TOE, by providing initial experimental evaluation results on popular low-power boards used in the TinyML community, for a wide variety of models from popular model zoo.


%



\section*{Acknowledgment}

The authors would like to thank Cedric Adjih and Nadjib Achir for useful discussions and suggestions. The research leading to these results partly received funding from the MESRI-BMBF German/French cybersecurity program under grant agreements No. ANR-20-CYAL-0005 and 16KIS1395K. The paper reflects only the authors’ views. MESRI and BMBF are not responsible for any use that may be made of the information it contains.

\ifCLASSOPTIONcaptionsoff
  \newpage
\fi



%

\printbibliography

@article{chen2018tvm,
  title={TVM: An automated end-to-end optimizing compiler for deep learning},
  author={Chen, Tianqi and Moreau, Thierry and Jiang, Ziheng and Zheng, Lianmin and Yan, Eddie and Cowan, Meghan and Shen, Haichen and Wang, Leyuan and Hu, Yuwei and Ceze, Luis and others},
  journal={arXiv preprint arXiv:1802.04799},
  year={2018}
}

@article{lima2019testbeds,
  title={Experimental environments for the Internet of Things: A review},
  author={Lima, Luis Eduardo and Kimura, Bruno Yuji Lino and Rosset, Val{\'e}rio},
  journal={IEEE Sensors Journal},
  volume={19},
  number={9},
  pages={3203--3211},
  year={2019},
  publisher={IEEE}
}

@ARTICLE{mitton2011testbeds,
  author={Gluhak, Alexander and Krco, Srdjan and Nati, Michele and Pfisterer, Dennis and Mitton, Nathalie and Razafindralambo, Tahiry},
  journal={IEEE Communications Magazine}, 
  title={A survey on facilities for experimental internet of things research}, 
  year={2011},
  volume={49},
  number={11},
  pages={58-67},
  doi={10.1109/MCOM.2011.6069710}}

@inproceedings{adjih2015iot-lab,
  title={FIT IoT-LAB: A large scale open experimental IoT testbed},
  author={Adjih, Cedric and Baccelli, Emmanuel and Fleury, Eric and Harter, Gaetan and Mitton, Nathalie and Noel, Thomas and Pissard-Gibollet, Roger and Saint-Marcel, Frederic and Schreiner, Guillaume and Vandaele, Julien and others},
  booktitle={2015 IEEE 2nd World Forum on Internet of Things (WF-IoT)},
  year={2015},
  organization={IEEE}
}

@article{baccelli2018riot,
  title={RIOT: An open source operating system for low-end embedded devices in the IoT},
  author={Baccelli, Emmanuel and G{\"u}ndo{\u{g}}an, Cenk and Hahm, Oliver and Kietzmann, Peter and Lenders, Martine S and Petersen, Hauke and Schleiser, Kaspar and Schmidt, Thomas C and W{\"a}hlisch, Matthias},
  journal={IEEE Internet of Things Journal},
  volume={5},
  number={6},
  
  year={2018},
  publisher={IEEE}
}

@misc{zephyr,
  title = {{Zephyr Operating System}},
  howpublished = {\url{https://www.zephyrproject.org/}},
  note = {Accessed: 2023-05-15}
}

@article{hahm2015operating,
  title={Operating systems for low-end devices in the internet of things: a survey},
  author={Hahm, Oliver and Baccelli, Emmanuel and Petersen, Hauke and Tsiftes, Nicolas},
  journal={IEEE Internet of Things Journal},
  volume={3},
  number={5},
  pages={720--734},
  year={2015},
  publisher={IEEE}
}

@article{banbury2021mlperf,
  title={Mlperf tiny benchmark},
  author={Banbury, Colby and Reddi, Vijay Janapa and Torelli, Peter and Holleman, Jeremy and Jeffries, Nat and Kiraly, Csaba and Montino, Pietro and Kanter, David and Ahmed, Sebastian and Pau, Danilo and others},
  journal={arXiv preprint arXiv:2106.07597},
  year={2021}
}

@article{banbury2020benchmarking,
  title={Benchmarking tinyml systems: Challenges and direction},
  author={Banbury, Colby R and Reddi, Vijay Janapa and Lam, Max and Fu, William and Fazel, Amin and Holleman, Jeremy and Huang, Xinyuan and Hurtado, Robert and Kanter, David and Lokhmotov, Anton and others},
  journal={arXiv preprint arXiv:2003.04821},
  year={2020}
}

@inproceedings{sudharsan2021tinyml-bench,
  title={Tinyml benchmark: Executing fully connected neural networks on commodity microcontrollers},
  author={Sudharsan, Bharath and Salerno, Simone and Nguyen, Duc-Duy and Yahya, Muhammad and Wahid, Abdul and Yadav, Piyush and Breslin, John G and Ali, Muhammad Intizar},
  booktitle={2021 IEEE 7th World Forum on Internet of Things (WF-IoT)},
  pages={883--884},
  year={2021},
  organization={IEEE}
}

@incollection{osman2022tinyml-bench,
  title={Tinyml platforms benchmarking},
  author={Osman, Anas and Abid, Usman and Gemma, Luca and Perotto, Matteo and Brunelli, Davide},
  booktitle={Applications in Electronics Pervading Industry, Environment and Society: APPLEPIES 2021},
  pages={139--148},
  year={2022},
  publisher={Springer}
}

@article{saha2022tinyml-review,
  title={Machine learning for microcontroller-class hardware-a review},
  author={Saha, Swapnil Sayan and Sandha, Sandeep Singh and Srivastava, Mani},
  journal={IEEE Sensors Journal},
  year={2022},
  publisher={IEEE}
}

@article{qiu2022ml-xray,
  title={ML-EXray: Visibility into ML deployment on the edge},
  author={Qiu, Hang and Vavelidou, Ioanna and Li, Jian and Pergament, Evgenya and Warden, Pete and Chinchali, Sandeep and Asgar, Zain and Katti, Sachin},
  journal={Proceedings of Machine Learning and Systems},
  volume={4},
  pages={337--351},
  year={2022}
}

@article{sze2017dnn,
  title={Efficient processing of deep neural networks: A tutorial and survey},
  author={Sze, Vivienne and Chen, Yu-Hsin and Yang, Tien-Ju and Emer, Joel S},
  journal={Proceedings of the IEEE},
  volume={105},
  number={12},
  pages={2295--2329},
  year={2017},
  publisher={Ieee}
}

@article{sanchez2020tinyml,
  title={Tinyml-enabled frugal smart objects: Challenges and opportunities},
  author={Sanchez-Iborra, Ramon and Skarmeta, Antonio F},
  journal={IEEE Circuits and Systems Magazine},
  volume={20},
  number={3},
  pages={4--18},
  year={2020},
  publisher={IEEE}
}

@article{ray2022tinyml-review,
  title={A review on TinyML: State-of-the-art and prospects},
  author={Ray, Partha Pratim},
  journal={Journal of King Saud University-Computer and Information Sciences},
  volume={34},
  number={4},
  pages={1595--1623},
  year={2022},
  publisher={Elsevier}
}

@Article{electronics10091012,
AUTHOR = {Sharma, Himanshu and Haque, Ahteshamul and Blaabjerg, Frede},
TITLE = {Machine Learning in Wireless Sensor Networks for Smart Cities: A Survey},
JOURNAL = {Electronics},
VOLUME = {10},
YEAR = {2021},
NUMBER = {9},
ARTICLE-NUMBER = {1012},
URL = {https://www.mdpi.com/2079-9292/10/9/1012},
ISSN = {2079-9292},
DOI = {10.3390/electronics10091012}
}

@inproceedings{2023_yousefzadeh-asl-miandoab_ProfilingMonitoringDeep,
  title = {Profiling and {{Monitoring Deep Learning Training Tasks}}},
  booktitle = {Proceedings of the 3rd {{Workshop}} on {{Machine Learning}} and {{Systems}}},
  author = {Yousefzadeh-Asl-Miandoab, Ehsan and Robroek, Ties and Tozun, Pinar},
  series = {{{EuroMLSys}} '23},
  doi = {10.1145/3578356.3592589},

}
%








\end{document}